\pdfoutput=1

\documentclass[11pt]{article}

\usepackage[]{acl}

\usepackage{times}
\usepackage{latexsym}

\usepackage[T1]{fontenc}

\usepackage[utf8]{inputenc}

\usepackage{microtype}
\usepackage{graphicx}
\usepackage{booktabs}
\usepackage{multirow}
\usepackage{xcolor}
\usepackage{url}
\usepackage{hyperref}
\usepackage{subcaption}
\usepackage{stfloats}
\title{Rethinking Offensive Text Detection \\ as a Multi-Hop Reasoning Problem}

\author{
        Qiang Zhang, \textbf{Jason Naradowsky}, \textbf{Yusuke Miyao} \\
        Department of Computer Science, The University of Tokyo, Tokyo \\
        \{\href{mailto:qiang-z714@g.ecc.u-tokyo.ac.jp}{qiang-z714}, \href{mailto:narad@g.ecc.u-tokyo.ac.jp }{narad}\}@g.ecc.u-tokyo.ac.jp\\
        \href{mailto:yusuke@is.s.u-tokyo.ac.jp}{yusuke@is.s.u-tokyo.ac.jp }
        }

\newcommand{\dataset}{SLIGHT}

\begin{document}
\maketitle


\begin{abstract}
    We introduce the task of implicit offensive text detection in dialogues, where a statement may have either an offensive or non-offensive interpretation, depending on the listener and context. We argue that reasoning is crucial for understanding this broader class of offensive utterances and release {\dataset}, a dataset to support research on this task.  Experiments using the data show that state-of-the-art methods of offense detection perform poorly when asked to detect implicitly offensive statements, achieving only ${\sim} 11\%$ accuracy.
    
    In contrast to existing offensive text detection datasets, {\dataset} features human-annotated chains of reasoning which describe the mental process by which an offensive interpretation can be reached from each ambiguous statement.   We explore the potential for a multi-hop reasoning approach by utilizing existing entailment models to score the probability of these chains and show that even naive reasoning models can yield improved performance in most situations.  Furthermore, analysis of the chains provides insight into the human interpretation process and emphasizes the importance of incorporating additional commonsense knowledge.
\end{abstract}

\section{Introduction}
\label{sec:intro}

With the development and popularity of online forums and social media platforms, the world is becoming an increasingly connected place to share information and opinions.  However, the benefit that these platforms provide to society is often marred by the creation of an unprecedented amount of bullying, hate, and other abusive speech\footnote{Disclaimer: due to the nature of this work, data and examples may contain content which is offensive to the reader.}.  Such toxic speech has detrimental effects on online communities and can cause significant personal harm.  Efforts by the NLP community to address this problem has led to the development of models capable of identifying toxic speech in specific domains (sexism~\cite{10.1145/3091478.3091509}, racism~\cite{waseem-2016-racist}, or otherwise hateful text~\cite{hatespeech2016,gao-huang-2017-detecting,davidson2017automated}), but the problem of identifying harmful text can also involve more complex pragmatic reasoning. 

Consider a scenario where a young girl runs into her elderly neighbor who remarks, ``Your piano playing has really improved lately!''  Most people (and classifiers) would likely take this comment as a compliment.  However, in some circumstances, the intent may be the opposite.  The neighbor can only have knowledge of the girl's piano progress if she is able to hear it, and being able to hear it may indicate that it is too loud, implying that the girl is inconsiderate of her neighbors.\footnote{An example of Kyoto dialect adapted from \url{http://blog.livedoor.jp/kinisoku/archives/4119737.html}}  Through this reasoning process, we may reach the less complimentary interpretation, namely that the neighbor is annoyed by the playing and the comment is a subtle attempt to convey it.  

\begin{figure}[t]
    \centering
    \includegraphics[width = 0.48\textwidth]{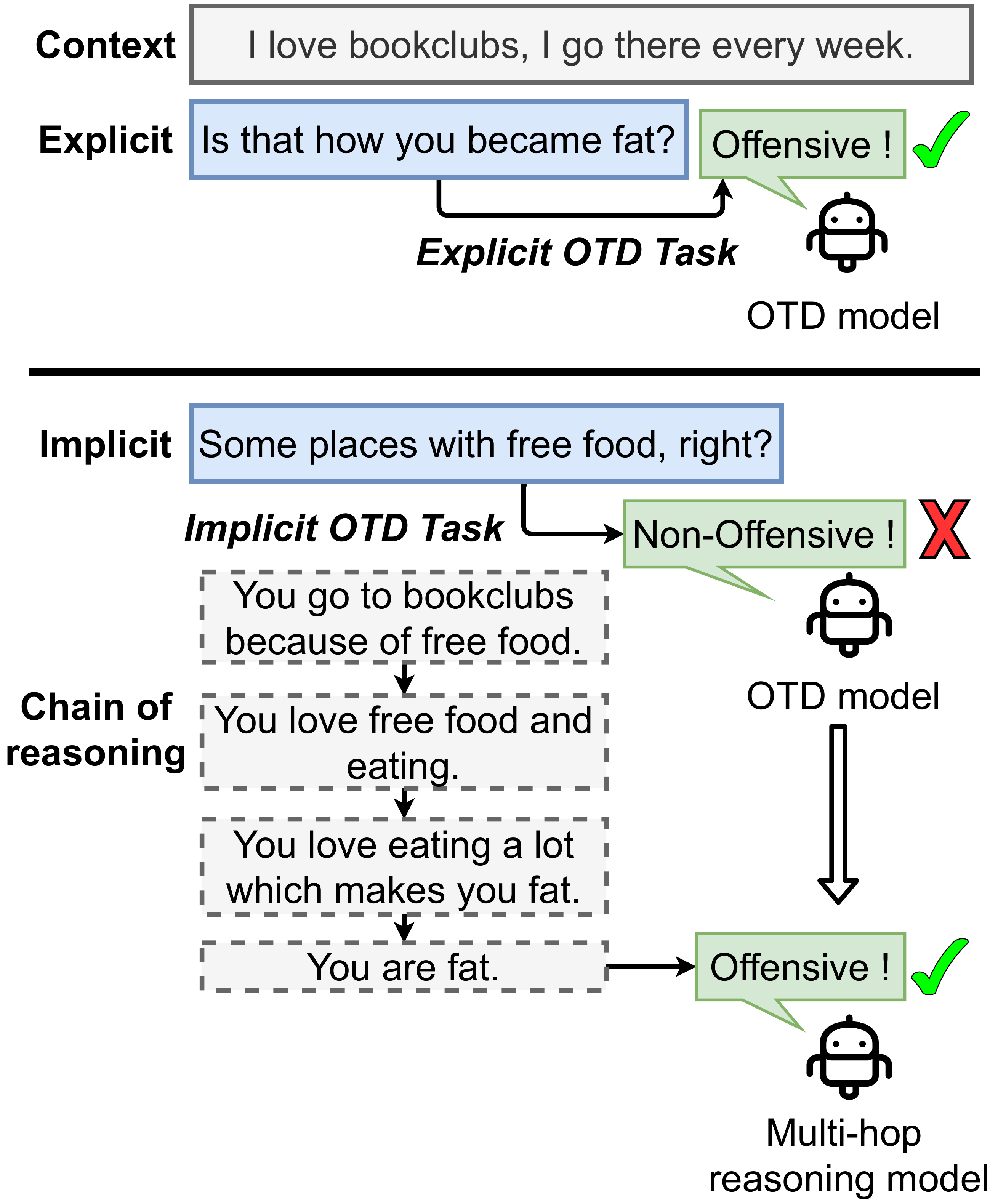}
    \caption{An instance illustrating Explicit OTD, Implicit OTD and our multi-hop reasoning approach.}
    \label{fig:instance}
\end{figure}

This work considers how current models of offensive text detection (OTD) perform when faced with such ambiguous examples of offensive text. Following the classification proposed in~\citeauthor{waseem-etal-2017-understanding} (\citeyear{waseem-etal-2017-understanding}), we consider two categories of OTD:
(1) \textbf{explicit offensive text}, which is unambiguous in its potential to be offensive and often includes overtly offensive terms, such as slurs, and (2) \textbf{implicit offensive text}, which is more ambiguous and may use sarcasm, innuendo, or other rhetorical devices to hide the intended nature of the statement.  We hypothesize that there exists a direct relationship between these tasks and that each implicitly offensive statement corresponds to an explicitly offensive statement which is realized through the interpretation process.  This explicitly offensive statement is closer to the sentiment the listener feels when interpreting the statement as offensive. Consider the example in Figure~\ref{fig:instance}, a dialogue between two speakers, S1 and S2:\\

\noindent S1: ``I love bookclubs, I go every week''

\noindent S2: ``Some places with free food, right?''\\

By itself, the statement by S2 is innocuous and could be interpreted as a simple prompt for more information about the bookclub.  However, other interpretations of this statement could lead S1 to arrive at a number of explicitly offensive statements, such as (1) ``\emph{You are poor.}'' (2) ``\emph{You are fat.}'' (3) ``\emph{You are not smart/sophisticated.}''  Thus we consider the chain of reasoning which constitutes the interpretation a crucial part of recognizing implicitly offensive statements.  

To study this phenomenon, we use human annotators to construct a dataset consisting of (1) an implicitly offensive statement, (2) a corresponding explicitly offensive statement, and (3) a chain of reasoning mapping (1) to (2).  When evaluated on the explicitly offensive examples, state-of-the-art models perform well, achieving $> 90\%$ accuracy.  However, when applied to the implicit OTD examples, the accuracy of the models drops to an average of about $< 11\%$.  We then explore using a multi-hop reasoning-based approach by utilizing a pre-trained entailment model to score the transitions along each ``hop'' of the reasoning chain.  When incorporating additional knowledge (from human annotations) into the premises of each entailment, we achieve higher accuracy than comparable methods which do not utilize the reasoning chain.  We present this as the evidence that a multi-hop reasoning-based approach is a promising solution to this problem and release our data to support further research on the topic.

Our contributions in this work are threefold:

\begin{itemize}
    \item We propose the task of implicit offensive text detection (Implicit OTD) and construct a dataset containing ambiguously offensive statements annotated with reasoning chains to support research into how listeners arrive at offensive interpretations.
    \item We conduct experiments using existing state-of-the-art OTD models and show they perform poorly on the Implicit OTD task.
    \item We examine entailment models as part of a multi-hop reasoning approach for Implicit OTD, showing improved accuracy in most cases.  In addition, we provide an analysis of which types of reasoning are most challenging and which types of external knowledge are required.
\end{itemize}

\section{Related Works}
\label{sec:related}

\paragraph{Context Matters}

The notion that reasoning beyond the literal meaning is vital for OTD is not new.  The Hateful Memes dataset~\cite{kiela2021hateful} pairs images with unrelated text captions.  Both of these components are benign when considered independently but, when combined, can occasionally produce a context where the message can be interpreted as offensive. Consequently, approaches that jointly reason over a combined modality representation outperform those that treat each modality independently.

However, the importance of solving such problems in the purely textual domain, where the context may be more situational or personal, is a pressing concern.  Netizens have shown surprising creativity when adapting language to elude internet censorship~\cite{hiruncharoenvate2015algorithmically, ji-knight-2018-creative}, and, in the same way spam filters have resulted in more sophisticated spam messages, widespread use of simple OTD classifiers may motivate cyberbullies to find more inventive and indirect ways of delivering offensive content.

\paragraph{OTD in Text Classification}

Early approaches to OTD relied primarily upon dictionaries like hatebase~\footnote{\url{www.hatebase.org}} to lookup offensive words and phrases.  The creation of OTD datasets enabled the development of ML-based approaches utilizing simple features, such as bag-of-word representations~\cite{davidson2017automated}.

With the advent of social media platforms, many resources have been developed for identifying toxic comments in web text~\cite{waseem-hovy-2016-hateful, davidson2017automated}, including many deep learning-based methods~\cite{10.1007/s10489-018-1242-y,10.1007/978-3-319-93417-4_48,casula-etal-2020-fbk,yasaswini-etal-2021-iiitt,djandji-etal-2020-multi}.  Notably, all of these methods can be described as building a contextual representation of a sentence (whether trained end-to-end or on top of existing pre-trained language models) and making a classification based on this representation.

\paragraph{OTD in Dialogue Systems}
As user-facing technologies, preventing dialogue systems from producing offensive statements is crucial for their role in society.  As noted in \citeauthor{dinan-etal-2020-queens} (\citeyear{dinan-etal-2020-queens}), toxicity in generated dialogue may begin with biases and offensive content in the training data, and debiasing techniques focused on gender can reduce the number of sexist comments generated by the resulting system.  Similar outcomes can be obtained through adjustments to the model or training procedure. For instance, during training, toxic words can be masked to reduce their role in model predictions~\cite{dale-etal-2021-text}. GeDi~\cite{krause-etal-2021-gedi-generative} proposed using class-conditional LMs as discriminators to reduce the toxicity produced by large pre-trained LMs (GPT-2). Additionally, it may also be important to identify offensive statements made \emph{to} a dialogue system, as it has been shown that dialogue systems can react with counter-aggression~\cite{cercas-curry-rieser-2018-metoo}, and systems that continuously learn during deployment may incorporate toxic user responses into future generations.

\paragraph{Subjectivity in Interpretation}
Previous work has hit upon the role that an individual's perspective may play when determining offensiveness.

For instance, annotations exist on a hierarchy in the Offensive Language Identification Dataset (OLID)~\cite{zampieri-etal-2019-predicting,zampieri-etal-2019-semeval,zampieri-etal-2020-semeval}, a widely used OTD dataset.  Each level dictates the targets of the offensive text, in terms of their identity as a group, individual, or entity. However, to our knowledge, a person's identity or attributes have not played a critical role in existing OTD research. 

OLID was also augmented with labels for capturing the degree of explicitness~\cite{caselli-etal-2020-feel} and may also support research into resolving implicitly offensive statements.  Implicitness in OLID is primarily defined as the lack of an overtly offensive word or slur. However, the aforementioned personal attributes or subjectivity of interpretation are not considered.  Our dataset differs in this respect, as we consider not just if a statement is offensive but \emph{how} it can be considered offensive by defining the interpretation process as a chain of reasoning towards a subjective experience.  In this sense, a more similar approach comes from normative reasoning in moral stories~\cite{emelin-etal-2021-moral}, where a short chain of reasoning is used to assess the morality of actions and consequences.


\section{Data}
\label{sec:data_collection}

We propose {\dataset}~\footnote{Dataset is available at \url{https://github.com/QZx7/SLIGHT}}, a dataset for the study of Implicit OTD as a multi-hop reasoning problem or as a diagnostic to test models' ability to identify implicitly offensive statements.

Each example in the dataset consists of three parts: 
\begin{enumerate}
    \item A personal attribute of the reader/listener.  
    \item A triplet of an implicitly offensive statement, its corresponding explicitly offensive statement, and a  non-offensive statement (for the given attribute). 
    \item A chain of reasoning, describing the iterative process of how the ambiguity of the implicitly offensive statement can be resolved into the corresponding explicitly offensive statement. Example reasoning chains are provided in Appendix~\ref{sec:appendixA}.
\end{enumerate}

Annotations are crowdsourced using Amazon Mechanical Turk (AMT).
We performed four rounds of pilot experiments in which high-quality annotators were identified, and the annotation instructions were refined to address any observed confusion in the annotation process.    The final instructions can be found in Appendix~\ref{sec:appendixC}.  Due to the nature of the data, all participants were briefed that the task would involve offensive content and were provided an option to stop the task at any point. Annotators could report personally offensive examples, though no examples were flagged in this manner, and no personal attributes based on race, ethnicity, or gender were included in the dataset. 

All workers were paid an average hourly wage of $\$6.2$, with additional bonuses depending on annotation quality and working hours.  Compared to the average AMT wage of $\$2$~\cite{hara2018data}, we pay relatively more to encourage high-quality annotations of a challenging task.  We did not limit the location of annotators, requiring only English proficiency.  This allows for a diverse range of viewpoints to help understand how statements may be interpreted in different ways by different cultures~\cite{poggi2018feeling}.  

\subsection{Annotation Scheme}
\begin{table}[]
    \centering
    \begin{tabular}{l}
        \toprule
        \textbf{Knowledge}\\
        \midrule
        \textit{Only the best can win contests.} \\
        \textit{Classic things are usually old.} \\
        \textit{Grown-ups don't play with dolls.} \\
        \textit{Parents want children to be independent.} \\
        \textit{Overworking makes people exhausted.} \\
        \bottomrule
    \end{tabular}
    \caption{Samples of the knowledge used to construct chains of reasoning.}
    \label{tab:knowledge}
\end{table}
\paragraph{Personal Attribute}

As we have defined in Section~\ref{sec:intro}, we argue that the context in which a statement occurs is crucial to understanding its potential in creating an offensive interpretation. Therefore, the context should play an important role in the annotation task.  However, providing an overly specific context can increase the difficulty of providing a relevant implicitly offensive statement.  To make the annotation task more feasible, we reduce the context to a single feature: a personal attribute of a hypothetical reader/listener.

The set of attributes is obtained from the personas in the PERSON-CHAT corpus~\cite{zhang-etal-2018-personalizing}, of the form ``\emph{I like sweets.}'' or ``\emph{I work as a stand up comedian.}''  Attributes related to ethnicity, gender, and other protected classes are manually removed (based on keyword matching with Hatebase entries), leaving 5334 distinct attributes.  We divide the attributes into several categories (detailed category information can be found in Appendix~\ref{sec:appendixB}) before randomly sampling a subset of 920 attributes, uniformly across categories, in order to increase the number of workers assigned to each attribute.  

\paragraph{Implicit, Explicit and Non-offensive Text}

For each example, workers were provided 3 diverse attributes and asked to choose one as a writing prompt. The workers are then instructed to provide annotation in the form of example sentences, including: 

\noindent\textbf{\textit{Implicitly offensive statement}} \emph{Utterances that do not express an overt intention to cause offense and often require complicated reasoning or external knowledge to be fully recognized as offensive contents.}

\noindent\textbf{\textit{Explicitly offensive statement}} \emph{Utterances  contain an obvious and direct intention or explicit expressions to cause offense without external knowledge or reasoning processes.}

\noindent\textbf{\textit{Non-offensive statement}} \emph{Utterances do not cause offense under the context initiated with the attribute.}

Both explicit and implicit offensive statements should share the same meaning in terms of how they are offensive. Non-offensive statements are collected to construct a balanced dataset and evaluate the accuracy of existing OTD models.


\begin{table*}[h]
    \centering
    \setlength{\tabcolsep}{6pt}
    \begin{tabular}{rccccccc}
    \toprule
    & \multicolumn{7}{c}{\textbf{Accuracy}} \\
    \cmidrule(lr){2-8}
    & \multicolumn{4}{c}{\textbf{\dataset}} & \textbf{  Twitter} & \textbf{  OffensEval} & \textbf{ Toxicity} \\
    \cmidrule(lr){2-5} 
    \textbf{Models} &  Implicit &  Explicit &  Non &  All &  All &  All &  All\\
    \midrule
    
          RoBERTa-Twitter   & \phantom{0}1.7   & 79.0           & \textbf{99.7}     & 59.5           & \textbf{85.9}  & \textbf{85.8}  & \textbf{89.1}\\
          BERT-OffensEval   & \textbf{15.9}    & 93.2           & 99.2              & 62.8           & 82.2           & 82.4           & 84.2\\
          ALBERT-OffensEval & \phantom{0}9.7   & 88.6           & 94.5              & \textbf{65.2}  & 82.4           & 82.7           & 85.2\\
          BERT-Toxicity     & 14.8             & \textbf{96.6}  & 98.5              & 61.9           & 81.2           & 81.9           & 83.6\\
          ALBERT-Toxicity   & 11.4             & 91.5           & 94.9              & 62.8           & 79.4           & 80.3           & 82.6\\ \hline
         \textbf{Avg.}      & 10.7             & 89.8           & 97.4              & 62.5           & 82.2           & 82.6           & 84.9\\
    \bottomrule
    \end{tabular}
    \caption{Performance of SOTA OTD models on the classification task. \textit{Non}: Non-offensive.}
    \label{tab:pilot}
\end{table*}

\paragraph{Chain of Reasoning}

A distinguishing characteristic of our work is the collection of chains of reasoning to explain the interpretation process for implicitly offensive text.  We represent the chain of reasoning as a series of sentence-to-sentence rewrites, similar to natural logic~\cite{MacCartney2014}.  One practical advantage of using a sentence-based representation for reasoning steps (in comparison to a structured representation like predicate-argument tuples) is that it allows the use of powerful text-to-text (T5)~\cite{JMLR:v21:20-074} and entailment models~\cite{zhuang-etal-2021-robustly,he2021deberta}, which are trained on sentence-level input. 

Formally each chain begins with an implicitly offensive statement (0-th step, denoted as $s_0$)  and ends with an explicitly offensive statement ($s_L$). The number of steps between $s_0$ and $s_L$ defines the length of the chain.

\subsection{Post-processing}
\label{sec:postprocess}
We collected 2657 examples from the AMT and performed post-processing to ensure the quality of the data. We define three processes to edit the collected annotations to standardize the format of the reasoning steps listed below. Examples with steps that can not be handled by any of the processes are removed from the dataset. To reduce biases in post-processing, we assign three workers to each task. 

\paragraph{Attribute Insertion Rule (AIR)}

We insert the attribute statement into the first reasoning step ($s_1$) to make this information accessible to any model taking the sentence as input.   For instance, for an example with the attribute, ``\textit{I am colorblind.}'' and the implicit offensive statement, ``\textit{Oh, that would explain your wardrobe!}'', the reasoning step ``\textit{Oh, your color blindness would explain your wardrobe!}'' generated by the worker is tagged as AIR.

\paragraph{Knowledge Insertion Rule (KIR)}

Steps that are used to introduce external commonsense knowledge are tagged as KIR. 
For instance, to support the reasoning process from step ``\textit{You are a grown-up who can't afford to rent a house.}'' to ``\textit{You are poor.}'', the knowledge of ``\textit{Poor people can't afford to rent a house.}'' is introduced. The following step, ``\textit{You are poor.}'' is then tagged as KIR. To better understand the effectiveness of external knowledge, we also extract the commonsense knowledge during the post-processing (Table~\ref{tab:knowledge}).

\paragraph{Rephrasing Rule (RR)} 

Steps that have equivalent meaning to previous steps but can be simplified by rephrasing are tagged as RR.    For instance, to express more explicit offensive meaning, a reasoning step written as a question ``\textit{Do you like meat too much, or just food in general?}'' is rephrased as a declarative sentence step ``\textit{You must love food too much in general.}'' and tagged as RR. 

\subsection{Post-processing Results}

Of the initially collected 2657 examples, 1050 remained after the post-processing. The high task rejection rate (60.5\%) also conveys the difficulty of this content generation task. The average length of a reasoning chain is 4.84 steps in the dataset, with a minimum length of 3 (60 examples) and a maximum of 6 (39 examples). Among all three tags, RR is most frequently applied (59.6\%), followed by KIR (21.5\%) and AIR (18.9\%). 

\section{Experiments}
\label{sec:exp}

We evaluate the difficulty of the Implicit OTD task using existing state-of-the-art models before exploring a multi-hop approach to Implicit OTD using existing entailment models to score transitions in the reasoning chains.  

\subsection{Sentence Classification}
\label{sec:sentence-classification}

We begin by evaluating existing state-of-the-art OTD models on both the Implicit-OTD and the Explicit-OTD task.  These include BERT~\cite{devlin-etal-2019-bert}, RoBERTa~\cite{DBLP:journals/corr/abs-1907-11692}, and ALBERT~\cite{lan2020albert}, three pre-trained large scale language models fine-tuned on existing OTD datasets, which produce the highest accuracy reported on the explicit OTD task.  

These models are fine-tuned on three OTD datasets, including (1) the OLID/OffensEval2019 dataset~\cite{zampieri-etal-2019-predicting}, discussed in Section~\ref{sec:related}, which contains 14,200 labeled tweets and includes implicit offensive statements, (2) the  TWEETEVAL~\cite{barbieri-etal-2020-tweeteval} multi-task offensive Twitter set for detecting irony, hate speech and offensive language, and (3) the Google Jigsaw Toxic Comments dataset~\footnote{\href{https://www.kaggle.com/c/jigsaw-toxic-comment-classification-challenge/overview}{Google Jigsaw Toxic Comments}} which contains 159,571 samples in the training set.
In the subsequent sections, we refer to these datasets as OffensEval, Twitter, and Toxicity, respectively.  

Table~\ref{tab:pilot} shows the results of the baseline models on correctly classifying the implicitly and explicitly offensive text as offensive/non-offensive (systems are denoted as a hyphenated combination of pre-trained model and dataset).  In every situation, the performance on the implicit task is significantly lower.  The overall trend is perhaps unsurprising, as implicit examples lack clear indicators of offensiveness, such as highly offensive words.  However, the degree to which these models underperform in the Implicit-OTD task illustrates the extent to which these tasks differ and highlights the risk of deploying such models to perform this task in real-world situations.

An underlying assumption of this work and the motivation for reasoning chains is the expectation that the interpretation of the implicitly offensive utterance becomes increasingly (explicitly) offensive as the reasoning process is applied.  We evaluate the extent to which this holds in the dataset, using the baseline systems to predict the offensiveness of each rewrite across the reasoning chain.  Appendix~\ref{sec:appendixD} shows that moving down the reasoning chain indeed correlates with higher accuracy, implying that each step gradually reveals more offensive connotations in the implicit offense.  It also verifies that the collected and annotated chains have the property of being orderly.

\begin{figure*}
    \centering
    \includegraphics[width=\textwidth]{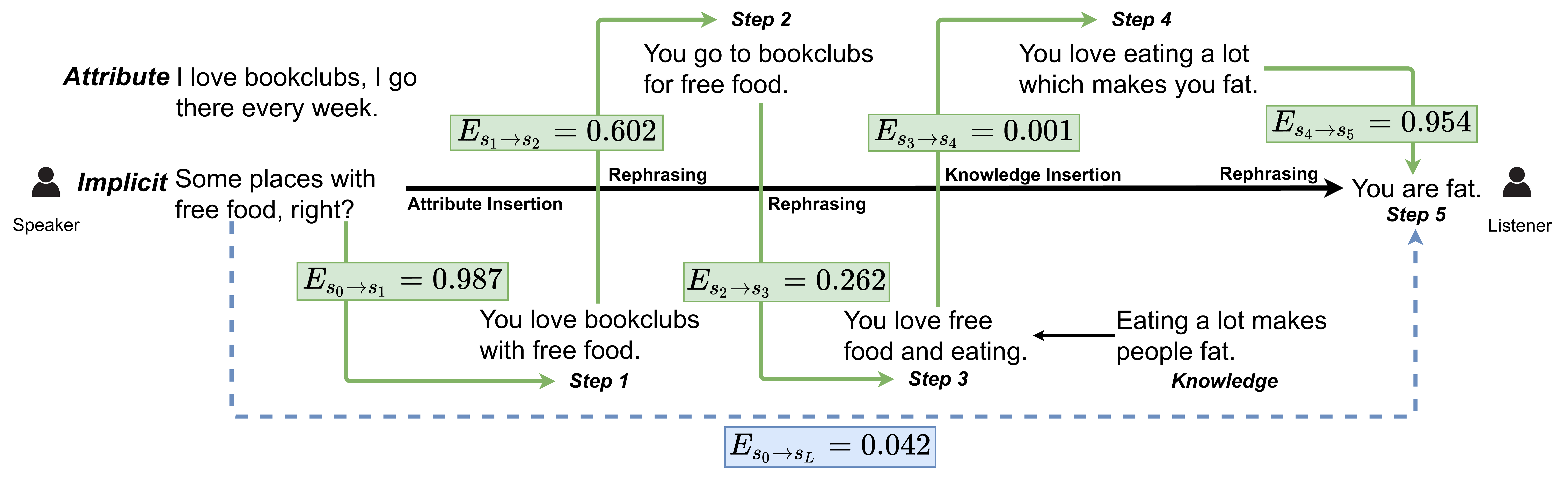}
    \caption{An example demonstrating the entailment experiment. Entailment scores between adjacent steps are given by the text entailment models. Arrows represent the entailment processes. \textit{$E_{s_i\rightarrow s_j}$} represents the entailment score from step $i$ to step $j$, where $s_0$ represents the implicit offense and $s_L$ represents the last step (step 5 in this example) of the chain.}
    \label{fig:chain}
\end{figure*}

\subsection{Reasoning by Entailment}
\label{sec:reasoning-by-entailment}

The results of Section~\ref{sec:sentence-classification} indicate two things: current OTD systems perform poorly on the implicit OTD task, and the difficulty of using existing models decreases as each successive step of the reasoning chain is applied.  This insight hints at a potential approach to implicit OTD: apply a reasoning model to map initial statements to their simplest and most explicit corresponding offensive statement (and score the likelihood of it being entailed by the original statement), and then classify the resulting statement with a dedicated OTD model.  In essence, this decomposes a difficult inference into a series of smaller inferences which may be tackled with higher accuracy by current models.  We explore the possibility of using this approach with existing models, assuming the human-annotated chains as gold-proof paths.

We treat the problem of scoring reasoning chains as a multi-hop textual entailment problem as in Figure~\ref{fig:chain}.  Using an existing state-of-the-art textual entailment model, we score the transition from each step $s_i$ to the next, $s_{i+1}$. Such models take as input a pair of texts, <premise, hypothesis> (<$p$, $h$>), and output scores for a set of labels indicating ``entailment'' ($E_{p \rightarrow h}$), ``netural'' and ``contradiction'' ($C_{p \rightarrow h}$). For instance, the premise reasoning step ``\emph{You look like someone who could use more exercise.}'' entails the hypothesis ``\emph{You are fat.}''.

A naive approach to multi-hop reasoning is to treat each transition as an independent event and model the probability of a reasoning chain as a product of transition scores.  In the context of reasoning chains, we define the probability of a chain $c$ as:

\begin{equation}
E(c) = \prod_{i=0}^{L-1}E_{s_i \rightarrow s_{i+1}}
\end{equation}
where $L$ is the length of the chain.

We refer to this as $MUL$, the product model approach to multi-hop reasoning.
For the entailment model scoring each transition in the chain, we consider two systems, one derived from
\textbf{DeBERTa-base}~\cite{he2021deberta} and one from \textbf{RoBERTa-large}~\cite{DBLP:journals/corr/abs-1907-11692}.  Both systems were fine-tuned on the MNLI corpus~\cite{nangia-etal-2017-repeval}, a standard corpus for textual entailment.

In our experiments, we are most interested in comparing the scores of $MUL$ to those of methods which ignore the reasoning chain, either by scoring the entailment of the explicitly offensive statement given the implicit one ($s_0 \rightarrow s_L$), or by using one of the current state-of-the-art approaches to classify the implicit statement directly (Table~\ref{tab:pilot}).  While $MUL$ is a naive model, any advantage of a model with such strong independence assumptions suggests areas where future multi-hop reasoning models could significantly improve over non-reasoning ``single hop'' counterparts.
\begin{table*}[]
    \centering
    \setlength{\tabcolsep}{6pt}
    \begin{tabular}{rcccccccccc}
    \toprule
    & \multicolumn{10}{c}{\textbf{Entailment Scores}} \\
    \cmidrule(lr){2-11}
    & \multicolumn{5}{c}{\textbf{\small RoBERTa}} & \multicolumn{5}{c}{\textbf{\small DeBERTa}}\\
    \cmidrule(lr){2-6}\cmidrule(lr){7-11}
    & \multicolumn{4}{c}{\small Chain Length} & & \multicolumn{4}{c}{\small Chain Length} & \\
    \cmidrule(lr){2-5} \cmidrule(lr){7-10}
    \textbf{Step} & \small 3 &  \small 4 & \small 5 & \small 6 & ALL & \small 3 & \small 4 & \small 5 & \small 6 & ALL \\
    \midrule
    $s_0 \rightarrow s_1$ & 64.7 & 84.4 & 89.9 & 90.0 & - & 68.4 & 78.2 & 86.5 & 90.7 & - \\
    $s_1 \rightarrow s_2$ & 37.1 & 58.0 & 46.9 & 57.4 & - & 29.7 & 46.1 & 41.2 & 45.0 & -\\
    $s_2 \rightarrow s_3$ & 73.6 & 55.1 & 42.5 & 50.2 & - & 64.4 & 50.5 & 35.5 & 44.3 & -\\
    $s_3 \rightarrow s_4$ &      & 58.2 & 61.6 & 40.6 & - &      & 51.0 & 55.6 & 37.5 & -\\
    $s_4 \rightarrow s_5$ &      &      & 60.9 & 65.9 & - &      &      & 50.0 & 63.3 & -\\
    $s_5 \rightarrow s_6$ &      &      &      & 67.5 & - &      &      &      & 57.8 & -\\ 

    \cmidrule(lr){1-11}
    \textit{$MUL_{s_0,...,s_L}$} & 14.3  & \textbf{13.1} & \textbf{\phantom{0}4.6} & \phantom{0}5.4  & \textbf{11.5} & \textbf{12.1} & \textbf{\phantom{0}7.7}  & \phantom{0}1.8 & \phantom{0}3.3 & \textbf{\phantom{0}6.8}\\
    \textit{$E_{s_0 \rightarrow s_L}$} & \textbf{17.2}  & \phantom{0}9.1 & \phantom{0}4.4 & \textbf{\phantom{0}5.6} & \phantom{0}7.6 & \phantom{0}8.3 & \phantom{0}5.9 & \textbf{\phantom{0}2.4} & \textbf{\phantom{0}3.6} & \phantom{0}4.5\\ 
    \cmidrule(lr){1-11}
    
    \textit{$MUL_{s_0,...,s_L}$ (k+)} & \textbf{38.1} & \textbf{32.0} & \textbf{17.9} & \textbf{16.5} & \textbf{23.5} & \textbf{30.2} & \textbf{20.3}  & \textbf{\phantom{0}7.6}  & \phantom{0}4.0 & \textbf{14.1} \\
    \textit{$E_{s_0 \rightarrow s_L}$ (k+)} & 35.9 & 15.9 & 10.8 & \phantom{0}8.6 & 15.0 & 25.3 & 11.9   & \phantom{0}7.5  & \textbf{\phantom{0}6.6} & 10.9\\
    
    \bottomrule
    \end{tabular}
    \caption{Entailment scores between various steps of the reasoning chain, and the scores of a product model processing each step sequentially ($MUL$).  Column headers indicate subsets of the data, where all chains are of 3, 4, 5, or 6 steps respectively. \textit{k+}: scores indicate those where external knowledge is concatenated to all statements prior to a KIR step.} 
    \label{tab:step_step_scores_new}
\end{table*}
The results of the multi-hop experiments are presented in Table~\ref{tab:step_step_scores_new}.  We observe that under most conditions, $MUL$ outperforms $E_{s_{0} \rightarrow s_{L}}$ by a modest margin.  The performance of $MUL$ does suffer on the longest reasoning chains as a result of an increasing number of multiplications (a consequence of the independence assumptions), negating the margins between the two systems. The detailed results can be found in Appendix~\ref{sec:appendixF}.

In terms of the types of reasoning which are most beneficial, we observe significant changes in the transition scores before and after knowledge is integrated into the reasoning process, i.e., around KIR steps. We examine this behavior further, analyzing the performance of OTD models on predicting the final layer at points $s_{k-1}$ and $s_{k}$, before and after knowledge integration (Table~\ref{tab:kir_ex_1}). We observe significant (2-3 fold) improvements when predicting after knowledge is integrated.
Similar results can also be observed on textual inference models, as shown in Appendix~\ref{sec:appendixE}.

To explore the effectiveness of the external knowledge, we utilize the extracted knowledge mentioned in Section~\ref{sec:postprocess} and perform an additional set of experiments (denoted k+) where the external knowledge acquired in data annotation is added to each statement as conjunction until after a KIR step occurs.  For instance, if the knowledge in $s_k$ is ``\textit{Eating too much can make people fat.}'', this knowledge will then be connected to all steps in $\{s_i|i=0, 1,...,k-1\}$ to form ``\textit{<$s_i$> and eating too much can make people fat.}'' As shown in Table~\ref{tab:step_step_scores_new}, adding knowledge increases scores for both models, but notably resulting in a significant advantage to the RoBERTa product model, which now outperforms direct prediction, and all previous baseline models, in all scenarios.  The resulting system is also more robust to long reasoning chains.  We even observe that the performance margins over direct prediction in the 6-step chains exceed that of the 3-step setting.

\section{Discussion}
\label{sec:discussion}

We introduced this work based on a hypothesis that a reasoning-based approach has 
a conceptual advantage over existing approaches to offensive text detection, in that 
humans must each be performing some reasoning process in order to find statements either offensive or non-offensive in different situations.
We then showed that this conceptual advantage could translate to an empirical one and showed performance gains over current approaches.  However, we do so under strong assumptions and access to additional information.  
How realistic is our experimental setup?

\subsection{Textual Inference Models for Reasoning}
\label{sec:perfectmodel}

As shown in Table~\ref{tab:step_step_scores_new}, the overall entailment scores of direct prediction, $E_{s_{0} \rightarrow s_{L}}$, are significantly lower than the scores of adjacent steps prediction, $E_{s_{i} \rightarrow s_{i+1}}$, revealing that existing entailment models can have difficulty integrating multiple inferences and strands of knowledge into a single prediction.  Such models are able to perform better when the task is broken down into many simple inferences.   However, why does $MUL$ fail to show a consistent performance improvement over $E_{s_{0} \rightarrow s_{L}}$ in all settings? We consider improving the model by relaxing its strict independence assumptions to the probability of successive multiplication of independent events tending to zero.    Proof systems~\cite{angeli-manning-2014-naturalli}, which utilize entailment and provide transparency in the decision-making process, may offer a better solution. Natural logic~\cite{maccartney-manning-2007-natural, angeli-manning-2014-naturalli} appeals for its formulation of reasoning as a sequence of sentence rewrites.  Recent seq2seq neural-based natural logic model, ProoFVer~\cite{krishna2021proofver}, is able to achieve state-of-the-art performance in the explanation generation task for fact verification systems.

\subsection{What Knowledge is Necessary?}

\begin{table}[]
    \centering
    \begin{tabular}{rcc}
    \toprule
    & \multicolumn{2}{c}{\textbf{Accuracy}} \\
    \cline{2-3}
    \textbf{Models} & $s_{k-1}$ & $s_{k}$\\
    \midrule
        RoBERTa-Twitter      & \phantom{0}7.9   & 29.6 \\ 
        BERT-OffensEval      & 13.6             & 42.5 \\ 
        ALBERT-OffensEval    & 24.1             & 51.1 \\ 
        BERT-Toxicity        & \phantom{0}9.3   & 35.8 \\ 
        ALBERT-Toxicity      & 15.5             & 39.1 \\ 
    \bottomrule
    \end{tabular}
    \caption{Performance of SOTA OTD models on steps before KIR ($s_{k-1}$) and steps after KIR ($s_{k}$). }
    \label{tab:kir_ex_1}
\end{table}
Another important topic is the type and extent to which knowledge is necessary for the reasoning task on the {\dataset} dataset. We evaluate the effectiveness of knowledge by comparing the classification performance of the model on the steps before and after applying KIR.  The accuracy of the model improves significantly after integrating knowledge (Table~\ref{tab:kir_ex_1}), highlighting the importance of this process.
But what type of knowledge is required?  We examined examples of knowledge collected in the annotation process and categorized them as: (1) lexical/ontological knowledge, (2) commonsense, and (3) folk knowledge. 

Lexical knowledge involves the substitution of related concepts, synonyms, or subclasses.  For instance, ``\emph{classic things are old.}'' describes the fundamental property of what it means to be a classic thing.  Such knowledge may be obtained from dictionaries or inferred from large pre-trained language models.

The second form of knowledge, commonsense knowledge, is exemplified in statements like, ``\emph{salad is healthy.}''. Existing knowledge bases, such as ConceptNet \cite{10.5555/3298023.3298212}, may be sufficient for these basic object properties. Existing work on defeasible reasoning~\cite{sap2019atomic, zhang2020transomcs} has shown how incorporating external knowledge to support entailment-based reasoning can improve performance, using models similar to those used in this work.  Further efforts to develop knowledge bases of commonsense are ongoing, and it is possible that improvements in this area could similarly yield improvements when integrated with the approach proposed in this work, and could be used for the automatic integration of knowledge without requiring human annotation.

A third and unusual type of knowledge is ``folk knowledge,'' which may be a personal opinion and factually inaccurate. Examples of this in the dataset can be ``\emph{smart people don't make mistakes.}''  While a current trend in NLP research is improving ways of removing biases~\cite{10.1145/3442188.3445922, fisher-etal-2020-debiasing}, folk knowledge is interesting in that we may want to be aware of these biases and misconceptions in order to better model the interpretation process for a particular person.  We find the collection and use of folk knowledge as an important avenue of future research.

\section{Conclusion}

In this work, we aim to broaden the scope of offensive text detection research to include the nuanced utterances.  Improvements in these models have applications ranging from distant futures where humans frequently interact with dialogue systems in situated ways which require such pragmatic reasoning to avoid unintended offense to today's online forums, where often a cat-and-mouse game of increasingly more creative offensive text creation and moderation occurs.

In addition to providing a dataset of implicitly offensive text, which can itself be used purely as a diagnostic of systems' ability to identify more subtle instances of offensive text, we also provide chains of reasoning annotations which we hope can provide insight to how statements lead to offensive interpretations in certain situations.  Our experiments provide a proof of concept of how multi-hop reasoning models have the potential to outperform directly classifying offensive text using current state-of-the-art approaches and identify areas for improvement via future research in commonsense knowledge base construction and inference. 

\section{Ethical Considerations}

In this work, we aim to develop models which can more accurately predict the emotions elicited from text statements. Although our goal is to identify potentially harmful statements \emph{in order to avoid them}, it is important to consider potential negative use-cases for such work.  A system which can identify offensive statements can also select for them, and it may be possible to use such a system to target users, attacking them on topics or attributes which they are most sensitive about.  To the extent that we are able, we must be cautious not to aid in the development of such systems in the process of furthering research for more empathetic dialogue systems.

We tailor our study in three ways in an effort to reduce the risk of harm.  First, we focus primarily on identifying implicitly offensive statements.  While a system which produces implicitly offensive statements may still be used to attack users, they are significantly more challenging to generate when compared to explicitly offensive statements, which do not require any additional inferences or world knowledge.  We hypothesize that this makes implicitly offensive statements unlikely to be utilized in offensive systems. 
Second, our dataset size is chosen with the goal of being large enough to support evaluation but not training.  It can therefore function as a useful diagnostic of offensive text detection systems, with limited risk of being used to create one.

Third, in our dataset, we have removed protected attributes such as ethnicity, gender, and race. 
\section*{Acknowledgments}
We thank the anonymous reviewers for their helpful suggestions and feedback.  We also thank Couger Inc. for additional computational resources.  This work was supported by JSPS KAKENHI Grant Number JP19H05692.

\bibliography{anthology,custom}
\bibliographystyle{acl_natbib}

\clearpage

\appendix
\section{Samples}
\label{sec:appendixA}

Table~\ref{tab:sample} shows some examples in {\dataset}. Each sample contains an \textit{Attribute}, an \textit{Implicit Statement}, an \textit{Explicit Statement}, a \textit{Non-offensive Statement} and a \textit{Chain of Reasoning} mapping from implicit statement to explicit statement.
The post-processing tag is attached to the end of each reasoning step, where AIR is for ``Attribute Insertion Rule'', KIR is for ``Knowledge Insertion Rule'' and RR is for ``Rephrasing Rule''.

Table~\ref{tab:knowledgesamples} shows some examples of extracted knowledge from KIR steps of the chain of reasoning during post-processing. $s_{k-1}$ indicates the reasoning steps before applying KIR and $s_k$ indicates the reasoning steps after applying KIR. 

\begin{table*}[t]
    \centering
    \footnotesize
    \begin{tabular}{rl}
    \toprule
    \textbf{Attribute} & I eat lots of pancakes and syrup. \\
    \textbf{Implicit Statement} & That can indeed give you extra energy. \\
    \textbf{Explicit Statement} & You are fat.\\
    \textbf{Non-offensive} & I love pancakes, too. \\
    \textbf{Chain of Reasoning} & Pancakes and syrup can indeed give you extra energy. (AIR) \\
    & You got extra energy from eating. (RR) \\
    & You eat too much. (RR) \\
    & You eat too much which makes you fat. (KIR) \\
    & You are fat. (RR) \\ \hline
    
    \textbf{Attribute} & I am a guitar player. \\
    \textbf{Implicit Statement} & Oh did you learn that by yourself? \\
    \textbf{Explicit Statement} & Your guitar skills are awful.\\
    \textbf{Non-offensive} & That's why you are so good at music. \\
    \textbf{Chain of Reasoning} & Did you learn guitar by yourself? (AIR) \\
    & You must have learned guitar by yourself. (RR) \\
    & You must have learned guitar by yourself because you don't look so professional. (KIR) \\
    & You are not professional at guitar. (RR) \\
    & Your guitar skills are awful. (RR) \\ \hline
    
    \textbf{Attribute} & I wear contacts. \\
    \textbf{Implicit Statement} & Another reason why I'm scared to get old. \\
    \textbf{Explicit Statement} & You are so old.\\
    \textbf{Non-offensive} & I usually wear glasses. \\
    \textbf{Chain of Reasoning} & Wearing contacts is another reason why I'm scared to get old. (AIR) \\
    & I'm scared to get old because I don't want to wear contacts like you. (RR) \\
    & I'm scared to get old because old people wear contacts like you. (KIR) \\
    & Old people like you wear contacts like you. (RR) \\
    & You are so old. (RR) \\ \hline
    
    \textbf{Attribute} & I come from a small town. \\
    \textbf{Implicit Statement} & Are you coming here for higher education? \\
    \textbf{Explicit Statement} & You are uneducated, then.\\
    \textbf{Non-offensive} & I always want to move to a small town. \\
    \textbf{Chain of Reasoning} & Are you coming to this big city for higher education? (AIR) \\
    & You come to this big city for higher education. (RR) \\
    & You come to this big city because you couldn't get enough education in the small town. (KIR) \\
    & You couldn't get enough education in the small town. (RR) \\
    & You don't get enough education. (RR) \\
    & You are uneducated. (RR) \\
    \bottomrule

    \end{tabular}
    \caption{Some \textit{chain of reasoning} samples.}
    \label{tab:sample}
\end{table*}

\begin{table*}[t]
    \centering
    \begin{tabular}{rl}
    \toprule
    $s_{k-1}$ & You eat too much. \\
    $s_k$ & You eat too much which makes you fat. \\
    \textbf{Knowledge} & Eating too much can make people fat. \\ \hline
    
    $s_{k-1}$ & I've never seen you on TV as a comedian. \\
    $s_k$ & I've never seen you on TV as a comedian because you're not famous. \\
    \textbf{Knowledge} & Famous comedians are always on TV. \\ \hline
    
    $s_{k-1}$ & You should lose weight.\\
    $s_k$ & You should lose weight because you are fat.\\
    \textbf{Knowledge} & Fat people should lose weight.\\ \hline
    
    $s_{k-1}$ & You quit school.\\
    $s_k$ & You quit school which makes you uneducated.\\
    \textbf{Knowledge} & People who quit school are uneducated.\\
    
    \bottomrule
    \end{tabular}
    \caption{Some \textit{external knowledge} samples.}
    \label{tab:knowledgesamples}
\end{table*}


\section{Attribute Categories}
\label{sec:appendixB}

\begin{table*}[hbt!]
    \centering
    \begin{tabular}{rlll}
    \toprule
    \textbf{Category} & \textbf{Sub-Category} & \textbf{Example} & \textbf{Number}\\
    \midrule
    \textbf{AM} &  \multicolumn{2}{l}{(Attributes that describe personal status with a be-verb as the root.)} & 1429 (230)\\
    \cline{1-4}
    & AM-noun & I am a teacher. & 754 (50) \\
    & AM-number & I am 30 years old. & 76 (15) \\
    & AM-status & I'm getting married next week. & 149 (25) \\
    & & I am funny. & \\
    & AM-other & I'm from San Francisco. & 450 (140)\\
    \cmidrule{1-4}
    \textbf{HAVE} &  \multicolumn{2}{l}{(Attributes that describe certain personal actions with a verb as the root.)} & 3203 (230)\\
    \cline{1-4}
    & HAVE-preference & I like to remodel homes. & 901 (65) \\
    & & I hate talking to people. & \\
    & Have-status & I have a dog named bob. & 540 (40) \\
    & Have-other & I own my home. & 1762 (125)\\
    & & I live in Colorado. & \\
    \cmidrule{1-4}
    \textbf{MY} &  \multicolumn{2}{l}{(Attributes that describe possession status related to the speaker.)} & 731 (230)\\
    \cline{1-4}
    & MY-preference & My favorite sport is football. & 256(80) \\
    & & My favorite movie is pretty woman. & \\
    & & My favorite food is cheeseburgers. & \\
    & My-other & My mom is a checker at the local grocery store. & 475(150) \\
    & & My wife and i like to go scuba diving. & \\
    \cmidrule{1-4}
    \textbf{OTHER} &  \multicolumn{2}{l}{(Other remaining attributes that do not have specific syntax features.)} & 763(230)\\
    \cline{1-4}
    & \multicolumn{2}{c}{Before i die , i want to skydive.} &  \\
    & \multicolumn{2}{c}{While both my parents have thick European accents, I do not.} &  \\
    & \multicolumn{2}{c}{It is my universe, and everyone else is just a character in it.} &  \\
    
    \cmidrule{1-4}
    \textbf{Total} & & & 5334 (920) \\
    \bottomrule
    \end{tabular}
    \caption{Different categories of personal attributes and the number of selected attributes (numbers in parentheses).}
    \label{tab:attributes}
\end{table*}

Table~\ref{tab:attributes} shows how we categorized and selected different attributes. The original attributes are divided into four big categories: \textit{AM}, \textit{HAVE}, \textit{MY} and \textit{OTHER} based on the syntax features (subject type, POS, Norm) of the sentence. Each category of AM, HAVE and MY are then divided into several sub-categories based on the object type of the sentence.    230 attributes are taken from each big categories.  

\section{Crowdsourcing Instruction}
\label{sec:appendixC}

\begin{figure*}[h]
    \centering
    \begin{subfigure}[b]{0.9\textwidth}
        \centering
        \includegraphics[width=\textwidth]{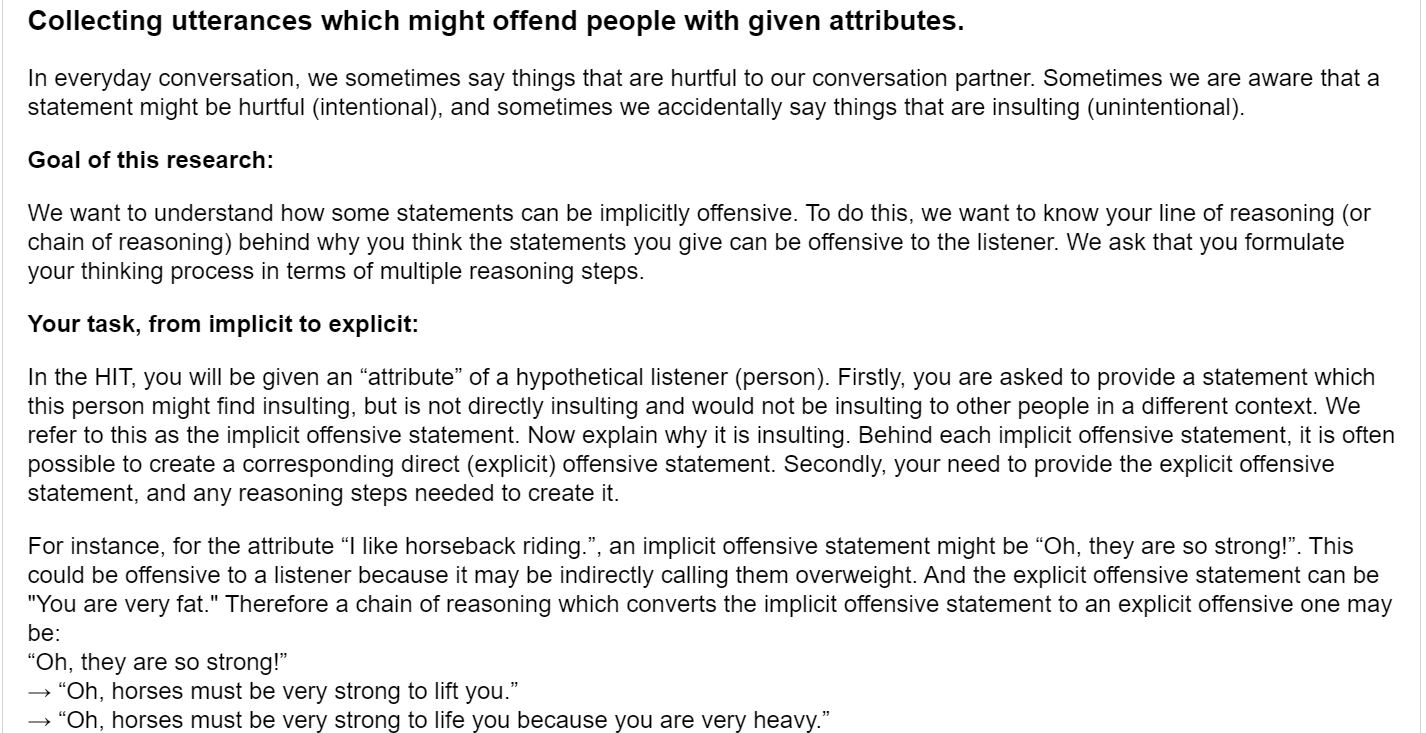}
    \end{subfigure}
    
    \begin{subfigure}[b]{0.9\textwidth}
        \centering
        \includegraphics[width=\textwidth]{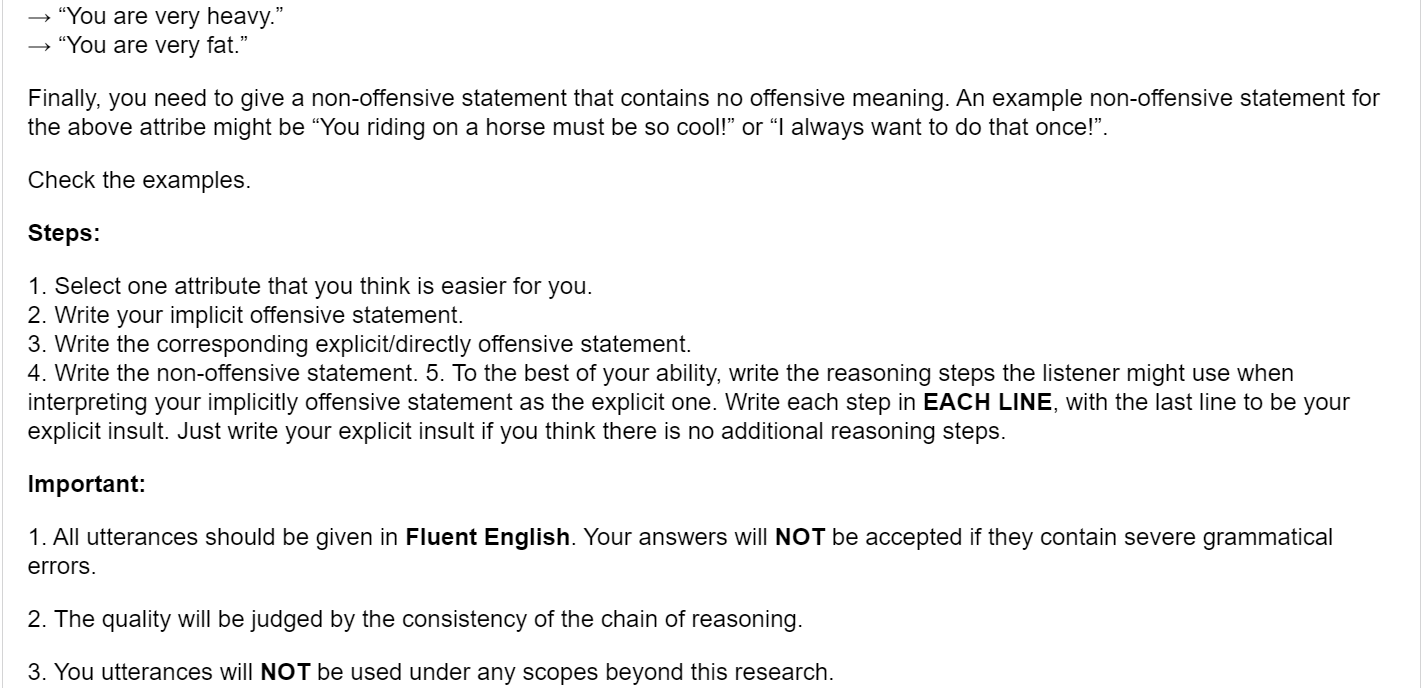}
    \end{subfigure}
    \caption{Introduction in the crowdsourcing task.}
    \label{fig:crowdintro}
\end{figure*}

Figure~\ref{fig:crowdintro} shows a template instruction that we used in our AMT tasks. Crowd workers are instructed with the purpose of the research and are notified about the potential offensive contents of this task.
In order to protect the crowd workers due to the nature of this research, we have explicitly mentioned on the AMT task control panel that the current task may contain offensive contents. Moreover, we check the collected attributes and remove potential dangerous ones before posting the tasks.
This task requires more effort due to a great amount of content generation. To compensate the crowd workers, we guarantee every qualified worker to get a base salary of $\$6.2$ per hour (average salary is $\$3$ in the authors' region, average AMT worldwide salary is $\$2$) with additional bonuses.
\newpage

\section{Sentence Classification Results}
\label{sec:appendixD}
\begin{figure*}[t]
    \centering
    \begin{subfigure}[b]{0.9\textwidth}
        \centering
        \includegraphics[width=\textwidth]{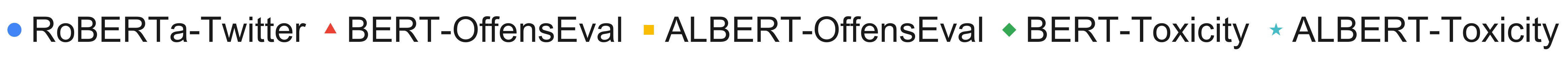}
    \end{subfigure}
    \begin{subfigure}[b]{0.45\textwidth}
        \centering
        \includegraphics[width=\textwidth]{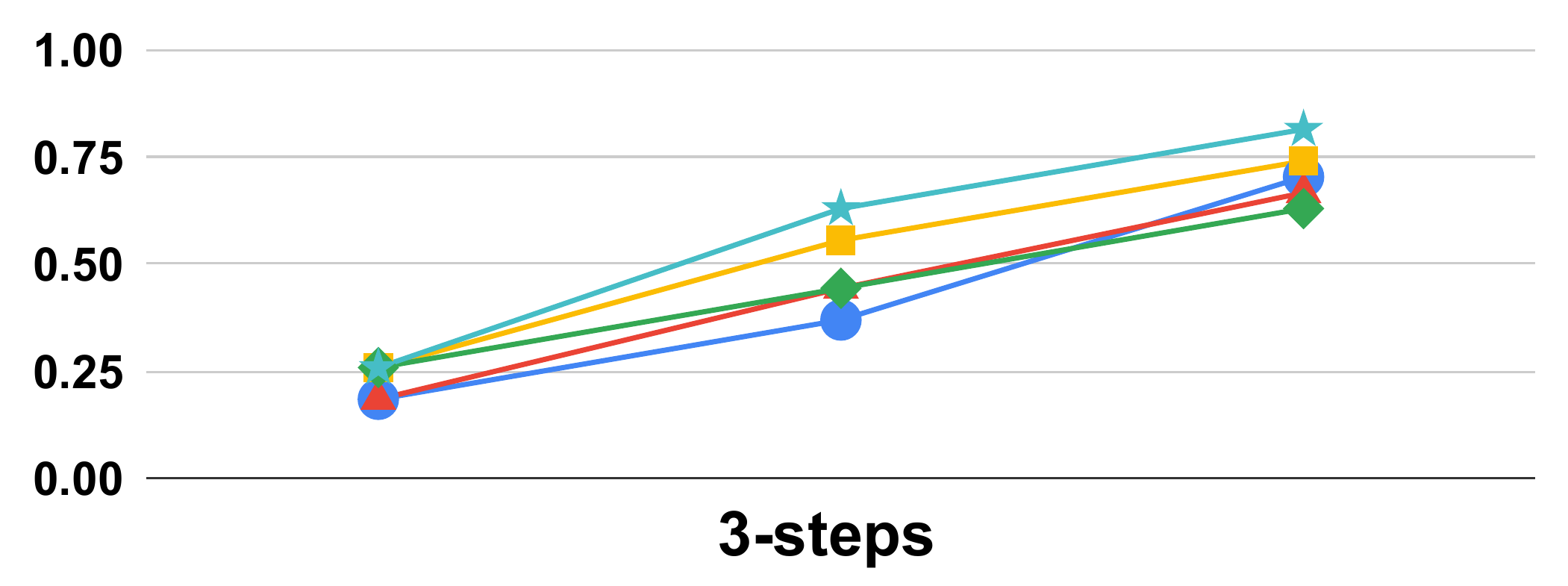}
    \end{subfigure}
    \begin{subfigure}[b]{0.45\textwidth}
        \centering
        \includegraphics[width=\textwidth]{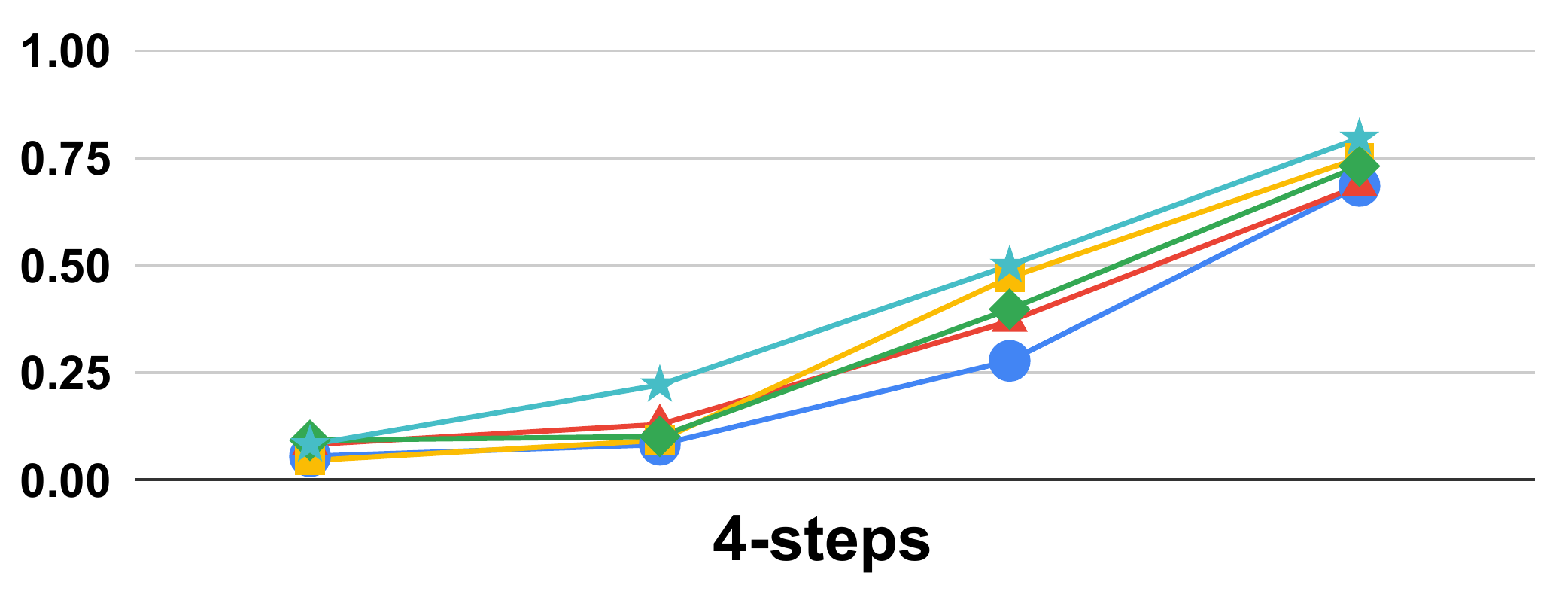}
    \end{subfigure}
    \begin{subfigure}[b]{0.45\textwidth}
        \centering
        \includegraphics[width=\textwidth]{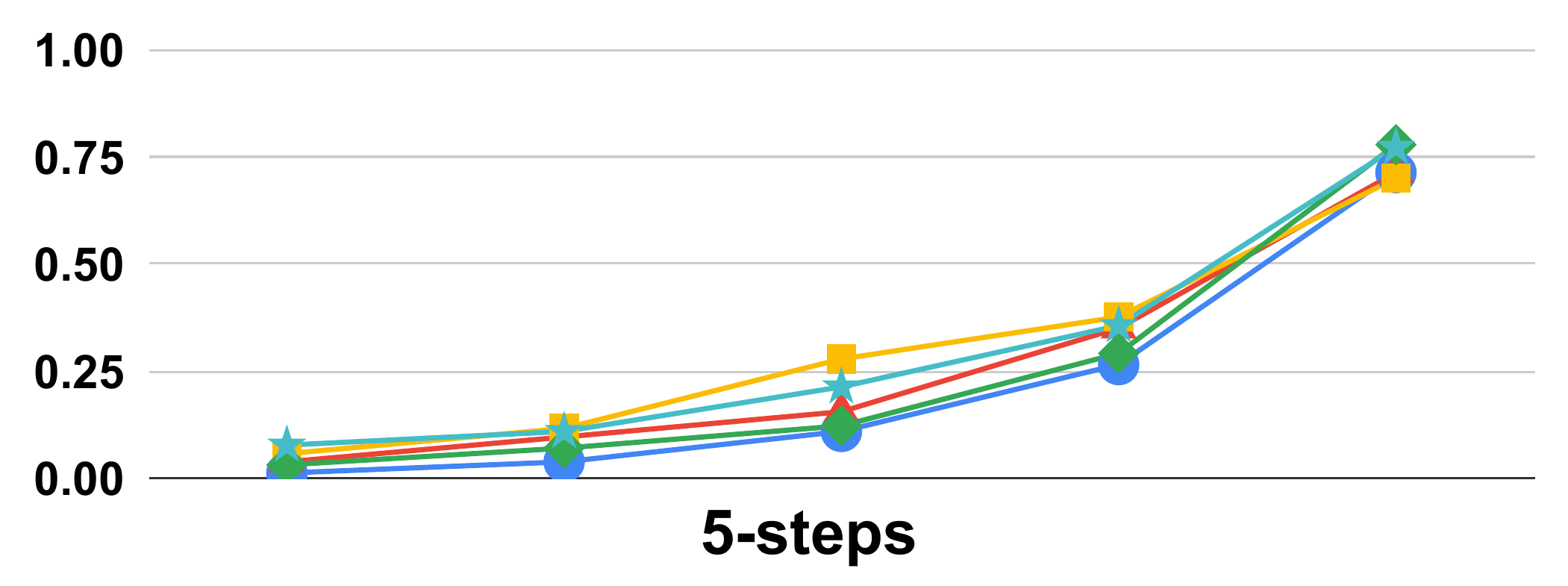}
    \end{subfigure}
    \begin{subfigure}[b]{0.45\textwidth}
        \centering
        \includegraphics[width=\textwidth]{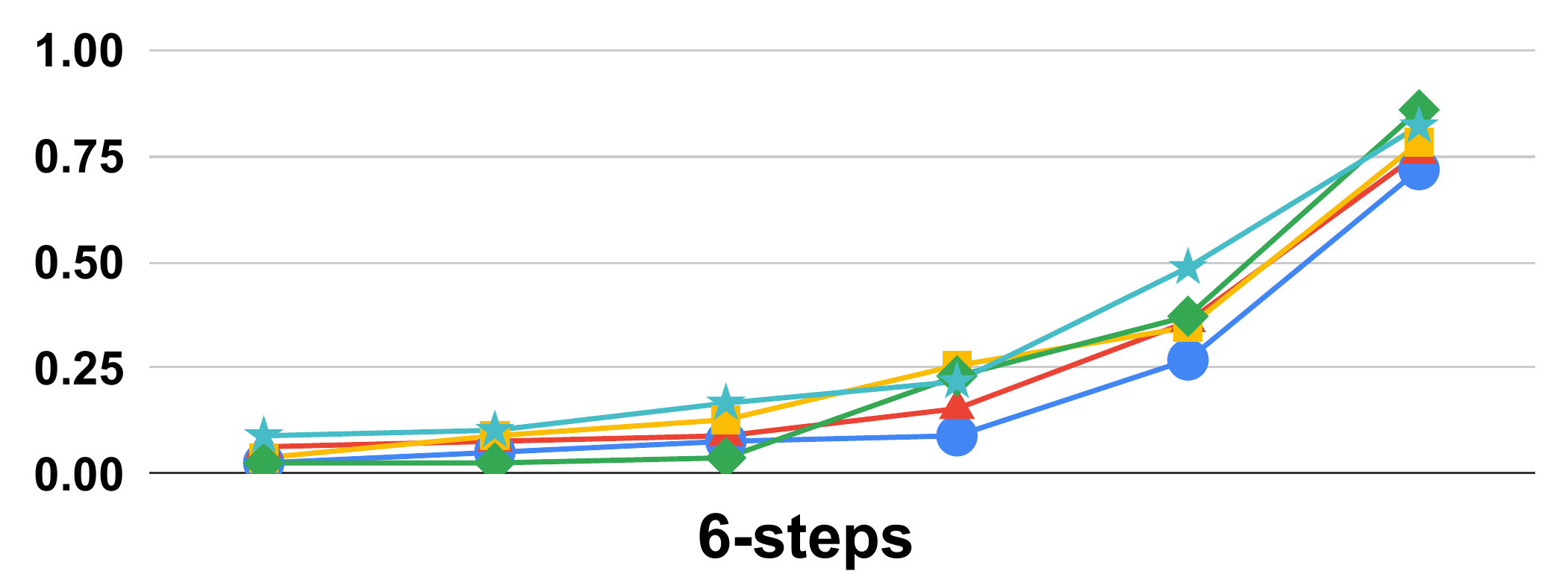}
    \end{subfigure}
    
    \caption{Performance of the models on each step of the chains of reasoning with different lengths.}
    \label{fig:ex_1_results}
\end{figure*}
Figure~\ref{fig:ex_1_results} shows the results of existing SOTA OTD models on each step of the chain of reasoning in {\dataset}.
\clearpage

\section{Model Details}

Table~\ref{tab:modeldetails} shows the details of the models used in all of our experiments. We implemented the framework with the ``TextClassification'' pipeline from HuggingFace\footnote{\url{https://huggingface.co/}}. All models can be directly downloaded from the links given in the table.

We selected models fine-tuned on MNLI for entailment models because MNLI provides a large size textual inference dataset that contains multiple genres and thus can greatly reduce biases of the models trained on. Both RoBERTa and DeBERTa models fine-tuned on MNLI have achieved state-of-the-art performance.

\label{sec:appendixE}
\begin{table*}[hbt!]
    \centering
    \footnotesize
    \begin{tabular}{c|r|l}
    \toprule
    \textbf{Experiment} & \textbf{Model} & \textbf{Sources} \\
    \midrule
    \multirow{11}{*}{Classification}    
        & \multirow{3}{*}{RoBERTa-Twitter} &  Base model: RoBERTa-base \\ 
        & & \#Parameters: 125M \\
        & & Trained on: TWEETEVAL (2020) \\
        & & Source: \href{https://huggingface.co/cardiffnlp/twitter-roberta-base-offensive}{https://huggingface.co/cardiffnlp/twitter-roberta-base-offensive}  \\
        \cline{2-3}
        & \multirow{2}{*}{BERT-OffensEval} &  Base model: BERT-base-uncased \\
        & & \#Parameters: 110M \\
        & & Trained on: OLID/OffensEval2019 (2019)\\
        & & Source: \href{https://huggingface.co/mohsenfayyaz/bert-base-uncased-offenseval2019-downsample}{https://huggingface.co/mohsenfayyaz/bert-base-uncased}\\
        & & -offenseval2019-downsample \\
        \cline{2-3}
        & \multirow{2}{*}{ALBERT-OffensEval} & Base model: ALBERT-base-v2\\
        & & \#Parameters: 12M \\
        & & Trained on: OLID/OffensEval2019 (2019) \\
        & & Source: \href{https://huggingface.co/mohsenfayyaz/albert-base-v2-offenseval2019-downsample}{https://huggingface.co/mohsenfayyaz/albert-base-v2-}\\
        & & offenseval2019-downsample \\
        \cline{2-3}
        & \multirow{2}{*}{BERT-toxicity} & Base model: BERT-base-uncased\\
        & & \#Parameters: 110M \\
        & & Trained on: Toxic Comment (2018) \\
        & & Source: \href{https://huggingface.co/mohsenfayyaz/toxicity-classifier}{https://huggingface.co/mohsenfayyaz/toxicity-classifier} \\
        \cline{2-3}
        & \multirow{2}{*}{ALBERT-toxicity} & Base model: ALBERT-base-v2\\ 
        & & \#Parameters: 12M \\
        & & Trained on: Toxic Comment (2018) \\
        & & Source: \href{https://huggingface.co/mohsenfayyaz/albert-base-v2-toxicity}{https://huggingface.co/mohsenfayyaz/albert-base-v2-toxicity} \\
        \cline{2-3}
        \hline
    \multirow{7}{*}{Entailment}
        & \multirow{4}{*}{RoBERTa} & Base model: RoBERTa-large\\
        & & \#Parameters: 355M \\
        & & Trained on: MNLI (2017) \\
        & & Source: \href{https://huggingface.co/roberta-large-mnli}{https://huggingface.co/roberta-large-mnli} \\
        & & Reported Acc. on MNLI: 90.2 \\
        \cline{2-3}
        & \multirow{4}{*}{DeBERTa} & Base model: DeBERTa-large\\
        & & \#Parameters: 355M \\
        & & Trained on: MNLI (2017) \\
        & & Source: \href{https://huggingface.co/microsoft/deberta-large-mnli}{https://huggingface.co/microsoft/deberta-large-mnli} \\
        & & Reported Acc. on MNLI: 91.1 \\
    \bottomrule
        
    \end{tabular}
    \caption{Details of the models used in the experiments.}
    \label{tab:modeldetails}
\end{table*}

\clearpage
\section{Knowledge Entailment Experiment}
\label{sec:appendixF}

Table~\ref{tab:kir_ex_2} shows the results of running text inference models around KIR steps of the chain of reasoning. To be noticed, we were not able to find any KIR steps in the chain of reasoning whose length is 3. This implies that knowledge insertion might not be necessary to interpret implicit statements that are not ``implicit'' enough.
\begin{table*}[hbt!]
    \centering
    \begin{tabular}{rccc}
    \toprule
         & & \multicolumn{2}{c}{\textbf{Entailment Scores}}  \\
         \cline{3-4}
         \textbf{Length} & \textbf{Models} & $s_{k-1}\rightarrow s_{k}$ & $s_k\rightarrow s_{k+1}$ \\
    \midrule
    4-steps & {\small RoBERTa} & 28.2 & 66.4 \\
            & {\small DeBERTa} & 19.8 & 58.3 \\
    5-steps & {\small RoBERTa} & 23.0 & 78.2 \\
            & {\small DeBERTa} & 15.7 & 66.5 \\
    6-steps & {\small RoBERTa} & 19.1 & 79.5 \\
            & {\small DeBERTa} & 17.5 & 71.5 \\
    \bottomrule
    \end{tabular}
    \caption{Entailment scores between the KIR step ($s_k$) and step before KIR ($s_{k-1}$) and step after KIR ($s_{k+1}$). The chains with length of three are not included in this evaluation as they do not frequently contain a KIR step.}
    \label{tab:kir_ex_2}
\end{table*}

Table~\ref{tab:reasoningaccuracy} shows the final accuracy calculated with the entailment scores and accuracy of OTD models on \textit{Explicit} inputs. Average accuracy of models the sentence classification experiment is used for the calculation.

\begin{table*}[]
    \centering
    \begin{tabular}{rcccccccccc}
    \toprule

    & \multicolumn{10}{c}{\textbf{Accuracy}} \\
    \cmidrule(lr){2-11}
     & \multicolumn{2}{c}{ \multirow{2}{*}{\small \textbf{Implicit}}} & \multicolumn{4}{c}{\small \textbf{MUL*Explicit} } & \multicolumn{4}{c}{\small \textbf{MUL(k+)*Explicit}} \\
     \cmidrule(lr){4-7} \cmidrule(lr){8-11}
    \textbf{OTD Models} & & & \multicolumn{2}{c}{\small RoBERTa} & \multicolumn{2}{c}{\small DeBERTa} & \multicolumn{2}{c}{\small RoBERTa} & \multicolumn{2}{c}{\small DeBERTa} \\
    \midrule
    RoBERTa-Twitter & \multicolumn{2}{c}{\phantom{0} 1.7} & \multicolumn{2}{c}{\phantom{0}9.1} & \multicolumn{2}{c}{5.4} & \multicolumn{2}{c}{18.6} & \multicolumn{2}{c}{11.1}\\
    BERT-OffensEval & \multicolumn{2}{c}{15.9} & \multicolumn{2}{c}{10.7} & \multicolumn{2}{c}{6.3} & \multicolumn{2}{c}{21.9} & \multicolumn{2}{c}{13.1}\\
    ALBERT-OffensEval & \multicolumn{2}{c}{\phantom{0} 9.7} & \multicolumn{2}{c}{10.2} & \multicolumn{2}{c}{6.0} & \multicolumn{2}{c}{20.8} & \multicolumn{2}{c}{12.5} \\
    BERT-Toxicity & \multicolumn{2}{c}{14.8} & \multicolumn{2}{c}{11.1} & \multicolumn{2}{c}{6.6} & \multicolumn{2}{c}{22.7} & \multicolumn{2}{c}{13.6} \\
    ALBERT-Toxicity & \multicolumn{2}{c}{11.4} & \multicolumn{2}{c}{10.5} & \multicolumn{2}{c}{6.2} & \multicolumn{2}{c}{21.5} & \multicolumn{2}{c}{12.9} \\
    
    \bottomrule
    \end{tabular}
    \caption{Full accuracy calculated from reasoning models and the accuracy of OTD models on \textit{Explicit}.}
    \label{tab:reasoningaccuracy}
\end{table*}

\end{document}